\documentclass{article}

\usepackage[final]{neurips_2019_ml4ps}

\usepackage[utf8]{inputenc} 
\usepackage[T1]{fontenc}    
\usepackage{hyperref}       
\usepackage{url}            
\usepackage{booktabs}       
\usepackage{amsfonts,amsmath}       
\usepackage{nicefrac}       
\usepackage{microtype}      
\usepackage{graphicx}
\usepackage{subcaption}

\usepackage{color}

\title{Using recurrent neural networks for nonlinear component computation in advection-dominated reduced-order models}

\author{%
  Romit Maulik \\
  Argonne Leadership Computing Facility\\
  Argonne National Laboratory\\
  Lemont, IL 60439 \\
  \texttt{rmaulik@anl.gov} \\
   \And
   Vishwas Rao  \\
   Mathematics and Computer Science Division \\
   Argonne National Laboratory\\
   Lemont, IL 60439 \\
   \texttt{vhebbur@anl.gov} \\
   \And
   Sandeep Madireddy  \\
   Mathematics and Computer Science Division \\
   Argonne National Laboratory\\
   Lemont, IL-60439 \\
   \texttt{smadireddy@anl.gov} \\
   \And
   Bethany Lusch  \\
   Argonne Leadership Computing Facility \\
   Argonne National Laboratory\\
   Lemont, IL 60439 \\
   \texttt{blusch@anl.gov} \\
   \And
   Prasanna Balaprakash  \\
  Mathematics and Computer Science Division \& \\
  Leadership Computing Facility\\
   Argonne National Laboratory\\
   Lemont, IL 60439 \\
   \texttt{pbalapra@anl.gov} \\
}

\begin{document}

\maketitle

\begin{abstract}
  Rapid simulations of advection-dominated problems are vital for multiple engineering and geophysical applications. In this paper, we present a long short-term memory neural network  to approximate the nonlinear component of the reduced-order model (ROM) of an advection-dominated partial differential equation. This is motivated by the fact that the nonlinear term is the most expensive component of a successful ROM. For our approach, we utilize a Galerkin projection to isolate the linear and the transient components of the dynamical system and then use discrete empirical interpolation to generate training data for supervised learning. We note that the numerical time-advancement and linear-term computation of the system ensure a greater preservation of physics than does a process that is fully modeled. Our results show that the proposed framework recovers transient dynamics accurately without nonlinear term computations in full-order space and represents a cost-effective alternative to solely equation-based ROMs. 
\end{abstract}

\section{Introduction}

High-fidelity simulations of systems characterized by nonlinear partial differential equations (PDEs) are computationally prohibitive for decision-making tasks in multiple engineering and geophysical applications. To address this issue, researchers have devoted  significant effort to developing reduced-order models (ROMs) of such systems with the aim of reducing the degrees of freedom of the forward problem to manageable magnitudes. Therefore, ROMs find extensive application in uncertainty quantification, control, and multifidelity optimization. The interested reader is directed to [1] and references therein for an excellent treatise on ROMs for engineering problems. A common ROM development procedure can be described by the following steps: (1) reduced basis identification, (2) system evolution in the reduced basis, and (3) reconstruction in full-order space for assessments.

In this paper, we utilize conventional ideas for reduced basis identification with the use of proper orthogonal decomposition (POD) [2] for finding the optimal global basis (i.e., step 1) and focus on using recurrent neural networks for efficient and accurate system evolution in reduced space (i.e., step 2). Our test case is given by the viscous Burgers equation formulated for a moving shock problem [3]. We note that this one-dimensional PDE possesses a quadratic nonlinearity and is frequently utilized as a prototype for assessing numerical methods  before their utilization in higher-dimensional phenomena. The governing PDE, initial and boundary conditions, and the analytical solution for this problem are given by 
\begin{align}
\label{Eq1}
\begin{gathered}
\dot{u}(x,t,\nu) + \mathcal{N}[u(x,t,\nu)] + \mathcal{L}[u(x,t,\nu); \nu] = 0, \quad (x,t,\nu) \in \Omega \times \mathcal{T} \times \mathcal{P}, \\
u(x, t) =\frac{\frac{x}{t+1}}{1+\sqrt{\frac{t+1}{t_{0}}} \exp \left(R e \frac{x^{2}}{4 t+4}\right)}, \\
u(x, 0) =\frac{x}{1+\sqrt{\frac{1}{t_{0}}} \exp \left(Re \frac{x^{2}}{4}\right)}, \ \  
u(0, t) =0, \ \ 
u(L, t) =0,
\end{gathered}
\end{align}
where $\nu = 1/Re$ is the viscosity and $\mathcal{N}$ and $\mathcal{L}$ correspond respectively to the quadratic nonlinearity and the linear operator in the viscous Burgers equation. In this study, we fix $\nu = 1e-3$, $L=1$, and the final time as $t_f = 2.0$ and seek to accelerate ROMs by performing a nonintrusive calculation of the nonlinear term $\mathcal{N}$, which leads to lower memory and compute cost requirements compared with traditional numerical techniques.


\section{POD-Galerkin projection and discrete empirical interpolation method}

The orthogonal nature of the POD bases may be leveraged for a Galerkin projection onto a linear subspace that hierarchically embeds information. We start by revisiting Equation (\ref{Eq1}) written in the form of a full-order evolution equation for fluctuation components using $N_f$ degrees of freedom:  
\begin{align}
\dot{\hat{\mathbf{u}}}_f(x,t,\nu) + \mathcal{N}_f[\hat{\mathbf{u}}_f(x,t,\nu)] + \mathcal{L}_f[\hat{\mathbf{u}}_f(x,t,\nu); \nu] = 0.
\end{align}
It can be expressed in the reduced basis as 
\begin{align}
\boldsymbol{\psi} \dot{\mathbf{a}_r}(t,\nu) + \mathcal{N}_f[\boldsymbol{\psi} \mathbf{a}_r(t,\nu)] + \mathcal{L}_f[\boldsymbol{\psi} \mathbf{a}_r(t,\nu); \nu] = 0,
\end{align}
where $\mathbf{a}_r : \mathcal{T} \times \mathcal{P} \rightarrow \mathbb{R}^{N_r}, \mathbf{a}_r  \in \alpha$, corresponds to the temporal coefficients at one time instant of the system evolution and $\boldsymbol{\psi} \in \mathbb{R}^{N_f \times N_r} $ represents the \emph{truncated} bases of the POD modes obtained from $N_s$ snapshots of the full-order solution having $\mathbb{R}^{N_f}$ degrees of freedom. The orthogonal nature of the reduced basis and the commutative property of the linear term can be leveraged to obtain
\begin{align}
\label{Eq4}
\dot{\mathbf{a}_r}(t,\nu) + \boldsymbol{\psi}^T \mathcal{N}_f[\boldsymbol{\psi} \mathbf{a}_r(t,\nu)] + \mathcal{L}_r[\mathbf{a}_r(t,\nu); \nu] = 0,
\end{align}
which we denote as the POD-Galerkin projection method (POD-GP). We remark that the calculation of $\mathcal{N}_f$ necessitates a re-projection of the reduced-order solution to full-order space and subsequent nonlinear term calculation throughout the domain. This process can be extremely expensive for advection-dominated problems such as ours. For comparison and for the generation of training data, we use the discrete empirical interpolation method (DEIM) [4], which reduces the number of nonlinear term calculations significantly. Our goal is to \emph{completely preclude} nonlinear term computation in full-order space and to utilize DEIM as a bridge between POD-GP and our formalism. We denote this approach as POD-ML. The time advancement of this hybrid system is then performed by using a simple first-order Euler integrator (although this may easily be extended to higher-order methods such as Runge-Kutta or Adams-Bashforth techniques).

The DEIM procedure is outlined in the following. Let $\boldsymbol{\varphi} \in \mathbb{R}^{N_f \times N_m}$ be the truncated POD basis matrix for snapshots of the nonlinear term $\mathcal{N}_f$, where $N_m$ is the number of vectors in the truncated basis. DEIM calculates a unit-vector matrix $\mathbf{P} \in \mathbb{R}^{N_f \times N_m}$, specifying $N_m$ locations (out of the $N_f$ locations in the full-order space), where $\mathcal{N}$ may be computed to construct an approximation for $\mathcal{N}_f$. Then we define a matrix $\mathcal{P} \in \mathbb{R}^{N_f \times N_f}$ as
\begin{align}
    \mathcal{P} = \boldsymbol{\varphi} (\mathbf{P}^T \boldsymbol{\varphi})^{-1} \mathbf{P}^T
\end{align}
and approximate Equation (\ref{Eq4}) as 
\begin{align}
\label{Eq5}
\dot{\mathbf{a}_r}(t,\nu) + \mathcal{L}_r[\mathbf{a}_r(t,\nu); \nu] + \boldsymbol{\psi}^{T} \mathcal{P} \mathcal{N}_f[\boldsymbol{\psi} \mathbf{a}_r(t,\nu)] = 0.
\end{align}
 
\begin{figure}
    \centering
    \begin{subfigure}{\textwidth}
      \centering
      \includegraphics[width=\textwidth]{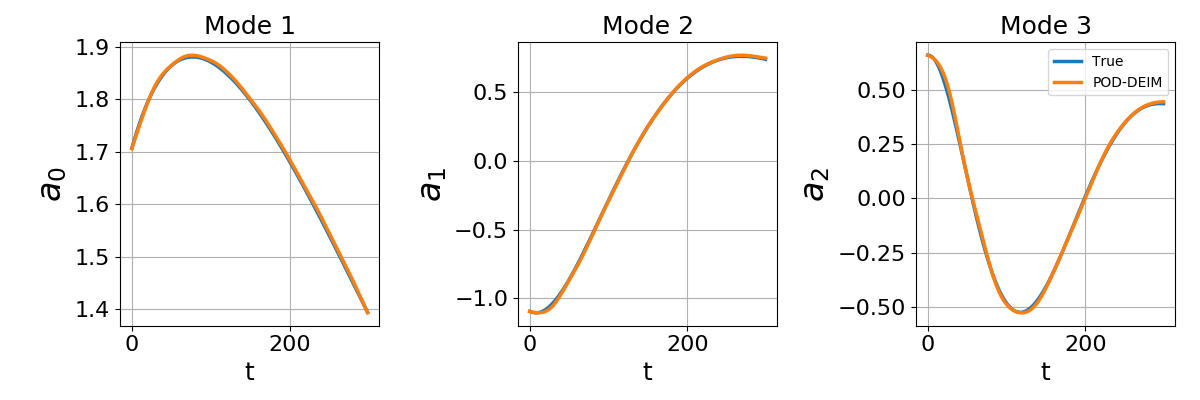}
      \caption{POD-GP}
      \label{Fig1:sfig1}
    \end{subfigure} \\ \vspace{-0.1cm}
    \begin{subfigure}{\textwidth}
      \centering
      \includegraphics[width=\textwidth]{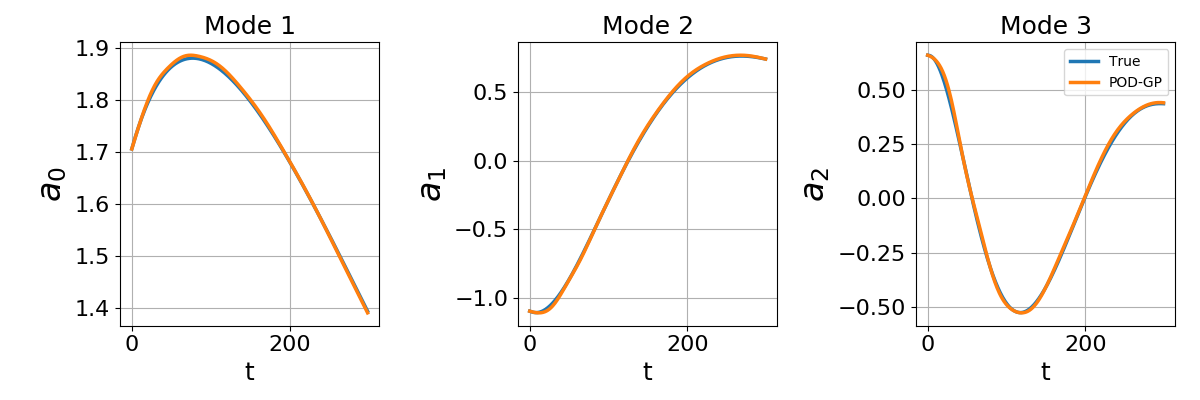}
      \caption{POD-DEIM}
      \label{Fig1:sfig2}
    \end{subfigure}
    \caption{Prediction utilizing 12 retained bases during truncation for the viscous Burgers equation. An identical performance validates the DEIM procedure.}
    \label{Fig1}
\end{figure}

Note that the linear operator and $\boldsymbol{\varphi} (\mathbf{P}^T \boldsymbol{\varphi})^{-1}$ are precomputed, leading to a reduction in the cost of calculating the nonlinear term  $\mathbf{P}^T \mathcal{N}_f$. Depending on $N_m$, this approach costs significantly less than POD-GP. The exact algorithm utilized to obtain $\mathbf{P}$ is a variant of least squares andis given  in [3]. Figure \ref{Fig1} shows a validation of our numerical method where both POD-GP and POD-DEIM are approximately equal in accuracy for our problem.

\section{Nonlinear surrogate using long short-term memory}

We utilized a long short-term memory (LSTM) neural network to further reduce the computational complexity of DEIM by calculating the nonlinear term, namely, $\boldsymbol{\psi}^{T} \mathcal{P} \mathcal{N}_h (t, \boldsymbol{\psi} \mathbf{a}_r)$ in Equation (\ref{Eq5}) nonintrusively. From this point, we denoted the DEIM approximation for the nonlinear term as $\mathcal{N}_r$. Our training data was generated by calculating DEIM coefficients from the analytical solution (keeping $N_r=12$ and $N_m=24$) for 300 snapshots of the nonlinear term in time.

We utilized a standard LSTM network [5] with one cell, $N_h=30$ hidden neurons, and a fully connected output layer to do many-to-one predictions for the nonlinear term $\mathcal{N}_r$ in a \emph{recursive fashion}. In other words, the outputs of the network are the DEIM nonlinear terms, which are reutilized as inputs for the next prediction. We used an input window of 10 to give an output of the nonlinear term at the next timestep, a batch size of 32, a learning rate of 0.001, and the ADAM optimizer. We observed that 60 epochs were sufficient for training the network under this setting. The training utilized a random 20\% validation split and saved the best model according to this data. Testing was performed through deployment within the hybrid numerical and ML framework as explained below. Training and assessment were performed by using an Intel core-i7 CPU with a basic build of TensorFlow 1.13.

\begin{figure}
    \centering
    \begin{subfigure}{\textwidth}
      \centering
      \includegraphics[width=\textwidth]{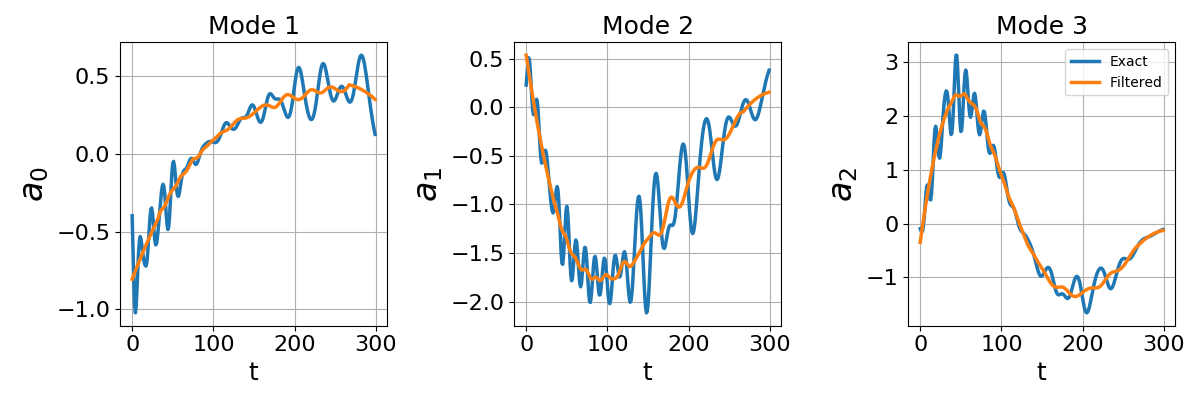}
      \caption{DEIM coefficient preprocessing for stability}
      \label{Fig2:sfig1}
    \end{subfigure} \\ \vspace{-0.1cm}
    \begin{subfigure}{\textwidth}
      \centering
      \includegraphics[width=\textwidth]{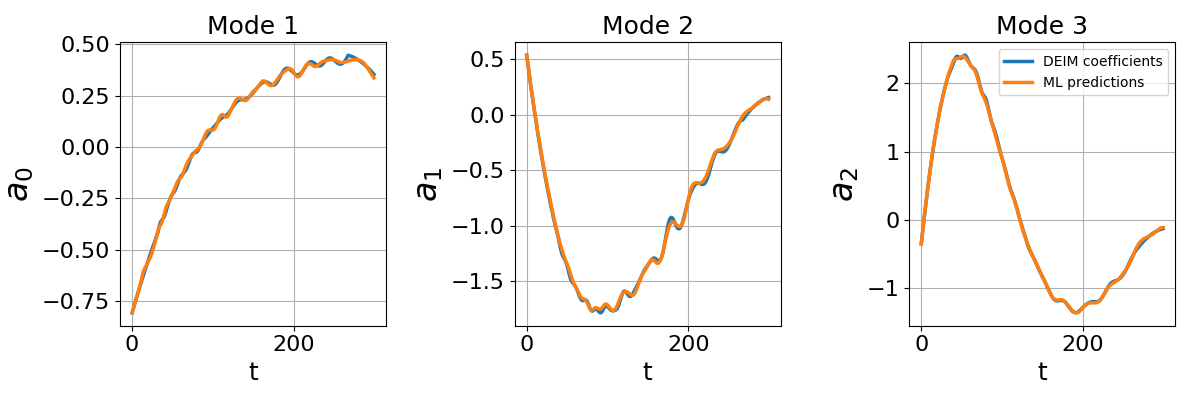}
      \caption{DEIM coefficient prediction by LSTM}
      \label{Fig2:sfig2}
    \end{subfigure} \\ \vspace{-0.1cm}
    \begin{subfigure}{\textwidth}
      \centering
      \includegraphics[width=\textwidth]{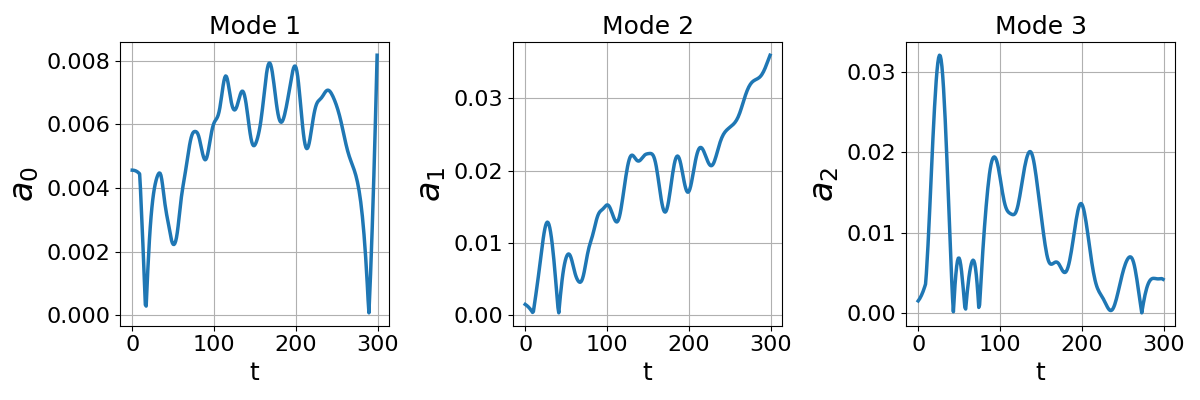}
      \caption{Relative error between POD-ML and POD-DEIM}
      \label{Fig2:sfig3}
    \end{subfigure}
    \caption{Preprocessing of DEIM coefficients (top), \emph{a priori} LSTM predictions for DEIM coefficients (middle) and relative errors between POD-ML and POD-DEIM (bottom) showing good agreement between the surrogate and equation based models.}
    \label{Fig2}
\end{figure}

Figure \ref{Fig2:sfig1} outlines the collected DEIM coefficients ($\mathcal{N}_r$) from the analytical solution showing oscillations. To predict these coefficients appropriately, we used a Savitsky-Golay low-pass spatial filter to mimic the effect of numerical smoothing in spatial and temporal discretizations. Figure \ref{Fig2:sfig2} shows the learning of the evolution of these coefficients (as an \emph{a priori} assessment). The LSTM predictions then were added to the linear component of the governing equation to obtain an accurate state advancement, as shown in Figure \ref{Fig2:sfig3}. Preprocessing of the oscillatory DEIM coefficients was crucial to the stability of the hybrid machine learning and numerical method. $L_2$-errors in modal evolution for each ROM technique (when compared withto the truth) were 0.064 for POD-GP, 0.059 for POD-DEIM, and 0.044 for POD-ML, indicating similar fidelity of the final solution. Assessments of computational cost are precluded here because for simple systems such as the Burgers equation, the cost of LSTM inference dominates the nonlinear computation. 
Benefits of the proposed method may be observed for larger systems with much greater degrees of freedom. For our problem, without any reduction, the per step complexity is $\mathcal{O}(N_f^2) \textrm{ FLOPS} + \mathcal{O}(N_f)$  nonlinear evaluations. POD-GP needs $\mathcal{O}(N_r^2) + \mathcal{O}(N_f N_r) \textrm{ FLOPS} + \mathcal{O}(N_f)$ evaluations, where $N_r \ll N_f$. POD-DEIM requires $\mathcal{O}(N_r^2) + \mathcal{O}(N_f N_r + N_f N_m) \textrm{ FLOPS}  + \mathcal{O}(N_m)$ nonlinear evaluations, where $N_m \ll N_f$. In contrast, our  proposed method utilizes $\mathcal{O}(N_h N_r + N_h N_h +N_r^2 +N_f N_r) \textrm{ FLOPS } $ and no online nonlinear evaluations. For larger problems where typically $N_h$, $N_m$, and $N_r$ are of the same order of magnitude, our method will yield exceptional gains.

\vspace{-0.3cm}

\section{Conclusions}

We have developed a method that precludes the reprojection of the ROM into full-order space for nonlinear term computation and, instead, utilizes prior knowledge of its evolution through the learning of precalculated DEIM coefficients. However, smoothing is required for the training data to mimic the stability-preserving nature of numerical methods. This requirement may also be true when the data comes from noisy physical or numerical experiments. The results here suggest that ROMs can be improved significantly for quick evaluation of extremely expensive physical systems with excellent retention of accuracy through the adoption of sequence learning techniques.

\subsubsection*{Acknowledgments}
This material is based upon work supported by the U.S. Department of Energy (DOE), Office of Science, Office of Advanced Scientific Computing Research, under Contract DE-AC02-06CH11357. This research was funded in part and used resources of the Argonne Leadership Computing Facility, which is a DOE Office of Science User Facility supported under Contract DE-AC02-06CH11357. This paper describes objective technical results and analysis. 


\section*{References}

[1] Taira, K., Brunton, S. L., Dawson, S. T., Rowley, C. W., Colonius, T., McKeon, B. J., \& Ukeiley, L. S. (2017). Modal analysis of fluid flows: An overview. {\it AIAA Journal}, 4013--4041.

[2] Maulik, R., Mohan, A., Lusch, B., Madireddy, S., Balaprakash, P \& Livescu D. (2019). Time-series learning of latent-space dynamics for reduced-order model closure. arXiv preprint arXiv:1906.07815.

[3] Berkooz, G., Holmes, P., \& Lumley, J. L. (1993). The proper orthogonal decomposition in the analysis of turbulent flows. {\it Annual Review of Fluid Mechanics}, \textbf{25(1)}, 539-575.

[4] Chaturantabut, S., \& Sorensen, D. C. (2010). Nonlinear model reduction via discrete empirical interpolation. {\it SIAM Journal on Scientific Computing}, \textbf{32(5)}, 2737--2764.

[5] Hochreiter, S., \& Schmidhuber, J. (1997). Long short-term memory. Neural Computation, \textbf{9(8)}, 1735--1780.

\begin{center}
    \framebox{\parbox{5in}{
    The submitted manuscript has been created by UChicago Argonne, LLC, Operator of Argonne National Laboratory (``Argonne''). Argonne, a U.S. Department of Energy Office of Science laboratory, is operated under Contract No. DE-AC02-06CH11357. The U.S. Government retains for itself, and others acting on its behalf, a paid-up nonexclusive, irrevocable worldwide license in said article to reproduce, prepare derivative works, distribute copies to the public, and perform publicly and display publicly, by or on behalf of the Government. The Department of Energy will provide public access to these results of federally sponsored research in accordance with the DOE Public Access Plan. \url{http://energy.gov/downloads/doe-public-access-plan}}}
    \normalsize
\end{center}

\end{document}